\documentclass[letterpaper, 10 pt, journal, twoside]{IEEEtran} 
\IEEEoverridecommandlockouts
\usepackage{graphicx}
\usepackage{setspace}          
\usepackage{url}
\usepackage{amsmath,bm}
\usepackage{upgreek}
\usepackage{algorithmic}
\usepackage{amssymb} 
\usepackage{mathtools}
\usepackage{xcolor} 
\usepackage{siunitx}
\usepackage{eucal}
\usepackage{systeme}
\usepackage[noadjust]{cite}
\usepackage{subfig}
\usepackage{array}
\usepackage{verbatim}
\usepackage{hyperref}
\hypersetup{hidelinks}
\usepackage{amsfonts}
\usepackage{microtype}
\usepackage{caption}
\usepackage{glossaries}
\usepackage{soul}
\usepackage{booktabs}
\usepackage{multirow}
\usepackage[all]{nowidow}
\newacronym{zmp}{ZMP}{Zero Moment Point} 
\newacronym{mpc}{MPC}{Model Predictive Control} 
\newacronym{srbd}{SRBD}{Single Rigid Body Dynamics}
\newacronym{grf}{GRF}{ground reaction force} 
\newacronym{com}{CoM}{center of mass} 
\newacronym{ud}{UD}{underdamped} 
\newacronym{cd}{CD}{critically damped} 
\newacronym{od}{OD}{overdamped} 
\newacronym{fft}{FFT}{Fast Fourier Transform}
\newacronym{rms}{RMS}{root-mean-square}
\newacronym{armpc}{ARMPC}{arm-aware MPC}
\newacronym{ocp}{OCP}{Optimal Control Problem}
\newacronym{lf}{LF}{Left Front}
\newacronym{lh}{LH}{Left Hind}
\newacronym{rf}{RF}{Right Front}
\newacronym{rh}{RH}{Right Hind}

\title{\LARGE \bf
ZiMPedance: Impedance-Aware ZMP Modeling and Control for Payload Carrying with Quadruped Robots
}
\author{Giovanni B. Dessy$^{1,2}$, Lorenzo Amatucci$^{1}$, Claudio Semini$^{1}$ and Victor Barasuol$^{1}$
\thanks{Manuscript received: June 10, 2026; Revised: September 3, 2026; Accepted: September 28, 2026.}%
\thanks{This paper was recommended for publication by Editor Cosimo Della Santina upon evaluation of the Associate Editor and Reviewers' comments.}%
\thanks{$^{1}$Dynamic Legged Systems Laboratory, Istituto Italiano di Tecnologia (IIT), Genova, Italy. E-mail: name.lastname@iit.it}%
\thanks{$^{2}$Universit\`a di Genova, Genova, Italy.}%
\thanks{Digital Object Identifier (DOI): see top of this page.}%
}

\begin{document}

\maketitle

\begin{abstract} 
Load transportation with quadruped robots is strongly affected by the
dynamics of the physical interface between the robot and the load.
Passive spring-based arms provide a lightweight and mechanically
compliant alternative to active manipulators, but their dynamics can
store and release energy during locomotion, producing oscillatory
interaction forces that reduce robot stability. This paper develops a passive-arm-aware ZMP model for a quadruped
carrying a payload through a passive arm. The reduced analytical model
relates passive-interface stiffness, damping, and payload mass to the ZMP
response under both external payload forcing and locomotion-induced base
excitation, showing that lightly damped passive modes can overlap with
gait harmonics. 
We then propose an arm-aware MPC (ARMPC), a passive-arm
Model Predictive Control approach based on extended Single Rigid Body
Dynamics that combines reduced floating-base dynamics with passive-arm
joint states, internal torsional springs and dampers, and base-arm
inertial coupling.
Simulation ablations show that ARMPC consistently reduces ZMP-margin
violations and prevents failures across all tested conditions, while
Nominal MPC, which neither models nor compensates the passive arm, and
Baseline MPC, which compensates it through an estimated wrench, exhibit
higher violation percentages and failures under disturbances.
Hardware pull-release experiments with a 2 kg payload show that the
proposed ARMPC completes all longitudinal and diagonal
disturbance trials without failure, whereas Baseline MPC fails in
7/15 and 8/15 trials, respectively.
The same model also enables
passive-arm end-effector tracking while maintaining a positive ZMP
margin.
Code is available at \url{https://github.com/iit-DLSLab/ZiMPedance}.
\end{abstract}
\section{INTRODUCTION}
\label{sec:introduction}

Heavy load transportation can expose human operators to significant risk
in logistics, industrial, and search-and-rescue scenarios. Legged robots
are well suited to assist with these tasks because they can traverse uneven
terrain~\cite{carryinguncarriableArashLab,CorosWheeledPlatforms} and carry
larger payloads than aerial platforms~\cite{quadrotorcarry,CoopOBJRot}.

This work considers a quadruped carrying a payload through a passive
spring-based arm. Compared with a rigid mount, the arm permits relative
payload motion and interaction with external agents, which is useful for
collaborative carrying and loco-manipulation.

Fully actuated manipulators provide greater dexterity and can render
compliant interaction through end-effector impedance control
~\cite{RisiglioneImpedanceControlQuadruped2022,
parosiKinematicallyDecoupledImpedanceControl2023}, but at the cost of
additional weight, complexity, and power. Passive spring--damper elements
provide compliance mechanically with lower actuation requirements, but
can store and release energy during locomotion, generating oscillatory
payload--base interaction forces. PACC~\cite{PACC_Paper} demonstrated
passive-arm payload carrying, but did not analyze how arm stiffness,
damping, and payload mass affect locomotion stability.

Payload transport with legged robots has been studied using different
physical interfaces, including cable towing
~\cite{yangCollaborativeNavigationManipulation2022}, ball-joint
connections~\cite{CentrvsDecKim}, and passive or pendulum-like links
~\cite{PACC_Paper,CASE_1dof_Arm}. Although mechanically different,
these interfaces introduce payload--base dynamics that can affect the
available stability margin.
Prior simulation results showed that gait dynamics and passive-arm
impedance can jointly affect the \gls{zmp} margin during payload carrying
~\cite{DessyCLAWAR2026}, motivating the impedance-aware stability
analysis developed here.
\begin{figure}[!t]
\centering
    \includegraphics[width=1\linewidth]{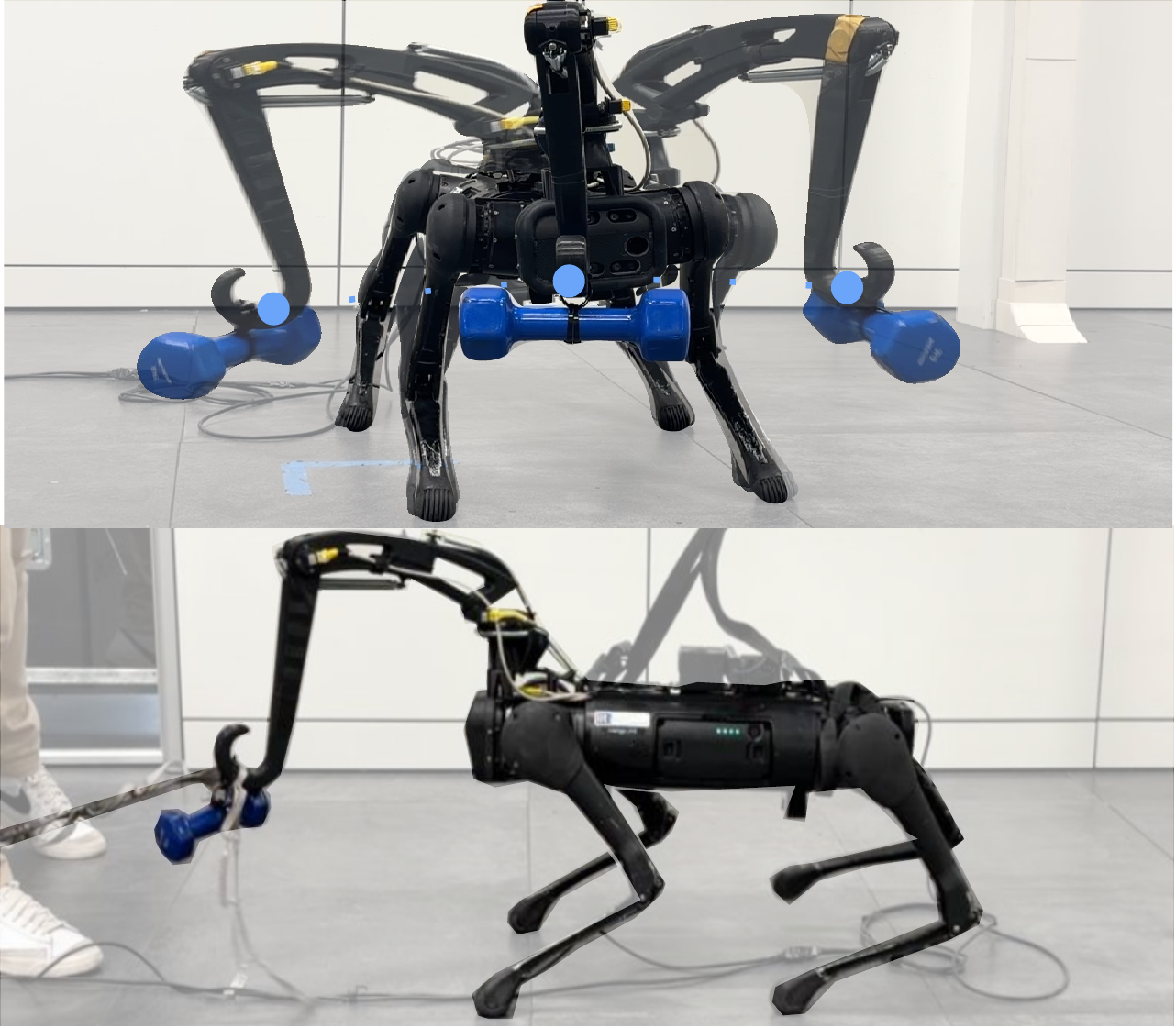}
    \caption{Top: Lateral end-effector tracking under a  $3\,\mathrm{kg}$ load. Bottom: pull-release disturbance with a $2\,\mathrm{kg}$ payload. }
    \label{fig:initial pic}
    \vspace{-0.3cm}
\end{figure}
Payload transport and loco-manipulation have also been addressed using
actuated arms~\cite{bellicosoALMAArticulatedLocomotion2019,
parosiKinematicallyDecoupledImpedanceControl2023,
dadiotisDynamicObjectGoal2025} and learning-based methods
~\cite{pandit2024learningdecentralizedmultibipedcontrol,
bethalaH2COMPACTHumanHumanoidCoManipulation2025,
anCollaborativeLocoManipulationPickandPlace2025,
pandit2026multiquadrupedcooperativeobjecttransport}.
These approaches achieve effective payload interaction through active
regulation or learned policies. In contrast, this work focuses on a passive
compliant interface and explicitly relates its impedance and carried mass
to locomotion stability.
\looseness=-1
Because passive-interface dynamics directly affect locomotion stability,
\gls{mpc} provides a natural framework in which stability and contact
requirements can be enforced over the prediction horizon. In quadruped
robotics, \gls{mpc} has been used for collaborative loco-manipulation
~\cite{vincentiCentralizedModelPredictive2023,
kimCooperativeLocomotionSupervisory2022}, while related carrying methods
incorporate interaction forces into gait generation, \gls{zmp} constraints,
or whole-body control~\cite{agravanteWalkingPatternGenerators2016a,
agravanteHumanHumanoidCollaborativeCarrying2019a,
zhongHumanoidRobotCooperative2021a,
hawleyControlFrameworkCooperative2019}. These approaches mainly consider
active or externally regulated interactions. Here, we focus on a passive
payload interface during low-speed crawl locomotion and use the
\gls{zmp}~\cite{ZEROMOMENTPOINTTHIRTY2004,
stairclimbzmp,bellicosoDynamicLocomotionOnline2018} to quantify its effect
on the stability margin.

The closest approach to the present work is the MPC controller used in
the paper that introduced the PACC mechanism~\cite{PACC_Paper}. 
Their controller includes the end-effector wrench as a
disturbance state whose horizontal force and yaw moment evolve only
through the spring deflection caused by base motion, with the
end-effector position assumed known; the remaining components are held
at their measured values. This quasi-static assumption holds for stiff,
well-damped arms, but neglects arm damping, inertia, and oscillation.
As shown in Secs.~\ref{sec:impedance_and_stability}
and~\ref{sec:section_violins}, these neglected dynamics are significant:
lightly damped passive modes are excited by the gait and reduce the
\gls{zmp} margin.
To account for these effects, this work introduces \gls{armpc}, which propagates the passive-arm states and
payload-induced motion inside the prediction model, in contrast to
~\cite{PACC_Paper}.

The main contributions of this work are:
\begin{itemize}
\item a reduced analytical model relating passive-arm impedance and
payload mass to the \gls{zmp} response under both external-force and
locomotion-induced base excitation;

\item a frequency-domain analysis showing how locomotion-induced base
excitation can overlap with passive-arm resonant modes and amplify the
resulting \gls{zmp} response;

\item \gls{armpc}, an extended \gls{srbd}-passive-arm
\gls{mpc} formulation that augments the reduced floating-base prediction model
with passive-arm joint states, torsional springs and dampers, and
base-arm inertial coupling, validated through simulation and hardware experiments.
\end{itemize}

\vspace{-0.2cm}
\section{BACKGROUND} 
\label{sec:background}
This section reviews the \gls{zmp} formulation used for stability analysis and introduces a mass-spring-damper abstraction of compliant
payload interfaces. This abstraction is used only in the reduced
analytical model of Sec.~\ref{sec:impedance_and_stability}; the \gls{mpc}
formulation in Sec.~\ref{sec:arMPC} uses the complete passive-arm
dynamics.
\subsection{The ZMP concept of stability}
\label{subsec:zmp_concept}

The \gls{zmp} is a well-known concept in legged robot control \cite{ZEROMOMENTPOINTTHIRTY2004}. It is defined as the point on the ground plane at which the net moment of all forces acting on the system has no component along the horizontal directions, assuming negligible centroidal angular momentum variations. Under standard assumptions, if the \gls{zmp} lies within the convex hull of the contact points, also known as the support polygon, the system is typically considered dynamically stable.

\begin{figure}[!h]
    \centering
    \includegraphics[width=1\linewidth]{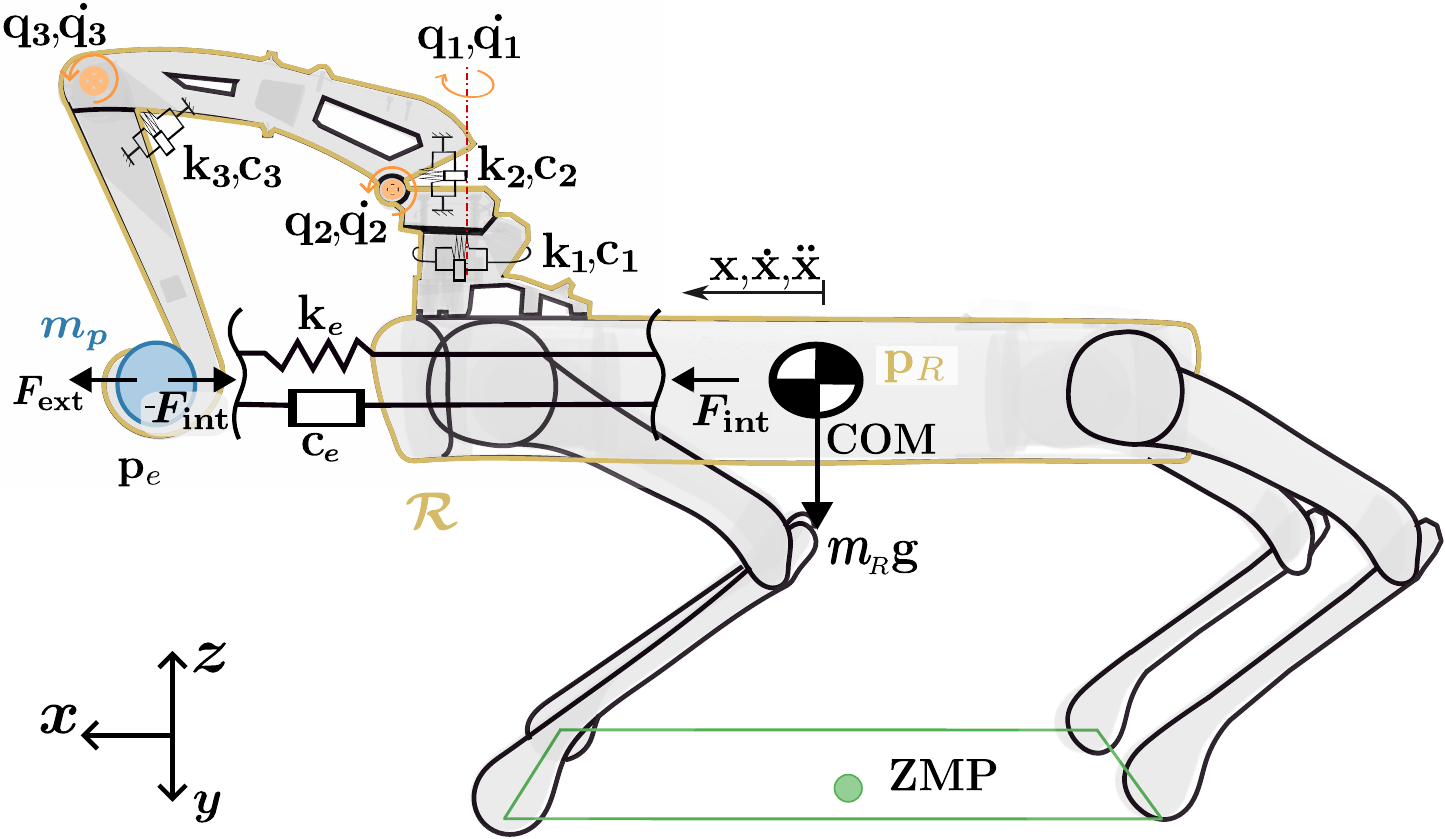}
    \caption{Passive arm and spring-damper interfaces on the quadruped.
    The support polygon, gravity vector, and end-effector interaction
    force are also shown.}
    \label{fig:Arm_views}
    \vspace{-0.25cm}
\end{figure}
To apply the standard ZMP formulation to the payload-carrying task, we
first define the free-body boundary used throughout the analysis. 
The free body, depicted in Fig.~\ref{fig:Arm_views}, contains the
quadruped and the complete passive arm, denoted by $\mathcal{R}$ and
outlined in yellow, while the carried payload is shown in light blue
outside this boundary.
Let $m_R$ be the mass of subsystem $\mathcal{R}$, and let
$\mathbf{p}_R=[x_R,y_R,z_R]^\top$ denote the position of the
\gls{com} associated with that subsystem.
The payload acts on the robot through the
interaction force
$\mathbf{f}_p=[f_{p,x},f_{p,y},f_{p,z}]^\top$
applied at the passive-arm end-effector position
$\mathbf{p}_e=[x_e,y_e,z_e]^\top$.
Consequently, the passive-joint spring and damping forces are internal
to $\mathcal{R}$ and do not appear independently in its ZMP balance.

We use an inertial world frame $\mathcal{W}$ with the $z$-axis pointing
upward, $\mathbf{g}=-g\mathbf{e}_z$ with $g>0$, and the $x$-axis
aligned with the nominal robot heading. Assuming negligible
centroidal-angular-momentum variation, the ZMP along the $x$ direction
on flat ground is

\begin{equation}
x_{\rm zmp}
=
\frac{
m_R x_R(g+\ddot{z}_R)
-
m_R z_R\ddot{x}_R
+
z_e f_{p,x}
-
x_e f_{p,z}
}{
m_R(g+\ddot{z}_R)-f_{p,z}
}.
\label{eq:zmp_external}
\end{equation}

For the low-speed carrying motions considered in the analytical model,
vertical CoM acceleration is assumed small,
$\ddot{z}_R\simeq0$. The corresponding expression along the $y$-axis
is obtained by replacing the $x$ components with their $y$
counterparts.
\vspace{-0.25cm}
    

\subsection{Passive Elements in Robotic Carrying Applications}
\begin{figure}[!ht]
    \centering
    \includegraphics[width=1\linewidth]{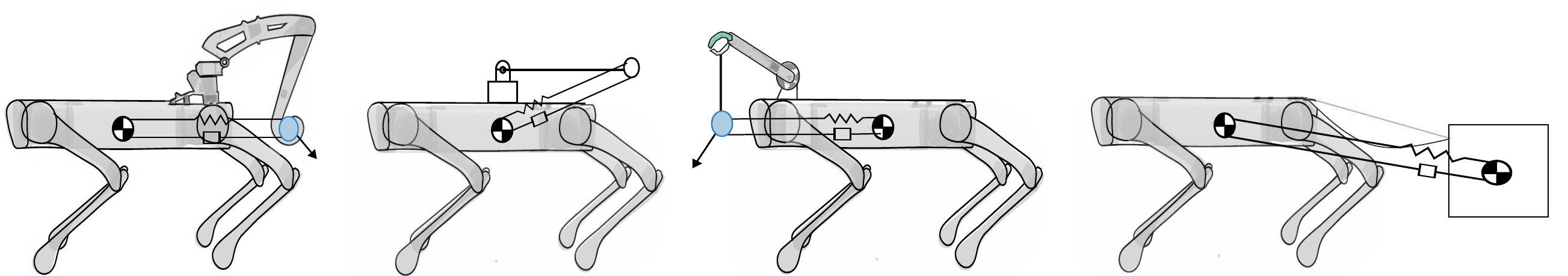}
    \caption{Examples of payload-interface mechanisms that can introduce passive interaction dynamics. From left to right: passive spring-based arm, ball-joint rigid inter-robot connection, one-DoF arm with pendulum-like payload behavior, and cable-towed payload.}
    \label{fig:payload_interfaces}
    \vspace{-0.4cm}

\end{figure}

Recent progress in payload-carrying with legged robots has led to
different mechanical interfaces between the robot and the load, some of
which are illustrated in Fig.~\ref{fig:payload_interfaces}. For the
reduced model analysis presented in Sec.~\ref{subsec:eq_model_passive_arm},
the interface-induced relative dynamics are locally approximated along
a selected direction by an equivalent mass-spring-damper model:
\begin{equation}
    F = m\ddot{x} + c\dot{x} + kx,
    \label{eq:msd}
\end{equation}
where $F$ is the generalized excitation force, $x$ is the relative
displacement across the payload interface, $m$ is the effective
payload-side moving mass, and $k$ and $c$ are the equivalent interface
stiffness and damping.
The mapping from the passive-arm joint-space impedance to this equivalent
one-DoF model is detailed next.
\subsection{Equivalent model of the passive arm}
\label{subsec:eq_model_passive_arm}

In this work, the mechanical interface is the three-DoF passive arm
introduced in~\cite{PACC_Paper} and illustrated in
Fig.~\ref{fig:Arm_views}. To connect this mechanism to the equivalent
one-DoF mass--spring--damper model in~\eqref{eq:msd}, let
$\mathbf q_a\in\mathbb{R}^3$ denote the passive-arm joint angles and
$\mathbf q_{a,0}$ their rest configuration. The passive-arm
spring-damper torques are
\begin{equation}
\boldsymbol{\tau}_s=-\mathbf K_q(\mathbf q_a-\mathbf q_{a,0}),
\qquad
\boldsymbol{\tau}_d=-\mathbf D_q\dot{\mathbf q}_a .
\label{eq:joint_impedance}
\end{equation}
where the diagonal matrices $\mathbf K_q$ and $\mathbf D_q$ contain the joint stiffness and
damping coefficients. By linearizing about the loaded equilibrium
$\mathbf q_{a,\mathrm{eq}}$, with
$\delta\mathbf p_e\simeq\mathbf J_{\mathrm{eq}}\delta\mathbf q_a$ and
$\mathbf J_{\mathrm{eq}}=\mathbf J_e(\mathbf q_{a,\mathrm{eq}})$, we obtain:
\begin{equation}
\mathbf K_x=
\mathbf J_{\mathrm{eq}}^{-\top}\mathbf K_q^{\mathrm{eq}}
\mathbf J_{\mathrm{eq}}^{-1},
\qquad
\mathbf D_x=
\mathbf J_{\mathrm{eq}}^{-\top}\mathbf D_q
\mathbf J_{\mathrm{eq}}^{-1},
\label{eq:cartesian_impedance}
\end{equation}
where
$\mathbf K_q^{\mathrm{eq}}=\mathbf K_q+\mathbf K_{g,p}$ is the
local stiffness about the loaded equilibrium, including the first-order
effect of payload gravity on the passive-joint restoring torque.
Projecting along a horizontal unit direction $\mathbf e$
yields:
\begin{equation}
k_e=\mathbf e^\top\mathbf K_x\mathbf e,
\qquad
c_e=\mathbf e^\top\mathbf D_x\mathbf e,
\qquad
m_e\simeq m_p ,
\label{eq:directional_impedance}
\end{equation}
where $m_e$ denotes the effective moving mass along the selected
direction. In the analytical derivation, we use the payload-dominated
approximation $m_e\simeq m_p$, neglecting the distributed inertia of the
passive-arm links. For the numerical frequency-response validation in
Sec.~\ref{sec:impedance_and_stability}, however, the scalar effective mass is refined to include the
projected inertial contribution of the passive-arm links while retaining
the same one-DoF reduced-model structure.
This scalar model is used only for the directional analysis. Cross-axis
coupling is neglected, and distributed arm-link inertia is omitted in the analytical approximation; the latter is included through
the effective moving mass used for frequency-response validation. The full
arm-link inertia and coupling are retained in simulation and in the ARMPC model.
\looseness=-1
To connect the passive-arm parameters to a frequency-domain analysis in
Sec.~\ref{sec:impedance_and_stability}, we use a local scalar approximation
dominated by the motion of the third passive joint. Let $x$ denote the
relative displacement across the payload interface, as in
\eqref{eq:msd}. Around the loaded equilibrium, $x\simeq l\,\delta q_{a,3}$,
where $\delta q_{a,3}$ is the displacement of the third passive joint
from its loaded-equilibrium value and $l$ is the effective moment arm
from that joint to the payload mass $m_p$:
\begin{equation}
c_e=\frac{d_3}{l^2},
\qquad
k_e=\frac{k_3}{l^2}+\frac{m_pg}{l},
\label{eq:pendulum_equivalent}
\end{equation}
where the complete three-joint equivalent stiffness and damping are
obtained from~\eqref{eq:cartesian_impedance}; the simplified expression in~\eqref{eq:pendulum_equivalent} isolates the
dominant pendulum-like contribution of the third link in the
reduced analytical model.
The corresponding natural frequency and damping ratio are:
\begin{equation}
\omega_n=\sqrt{k_e/m_p},\qquad
\zeta=c_e/(2\sqrt{m_p k_e}).
\label{eq:damping_ratio}
\end{equation}
The cases $\zeta<1$, $\zeta=1$, and $\zeta>1$ define \gls{ud}, \gls{cd} and \gls{od} regimes used in Sec.~\ref{sec:impedance_and_stability} and Sec. ~\ref{sec:section_violins}.

\section{Passive-Arm Effects on ZMP Stability}
\label{sec:impedance_and_stability}

\subsection{Base-Excited Passive-Arm ZMP Model}
\label{sec:zmp_impedance_model}

Using the equivalent stiffness, damping, and moving-mass parameters
introduced in Sec.~\ref{subsec:eq_model_passive_arm}, we relate the
passive arm-payload dynamics to the \gls{zmp} along the longitudinal
direction. The same scalar construction can be applied to the lateral
direction by replacing the $x$-axis quantities with their $y$-axis
counterparts.
Let $x_b$ and $x_p$ denote the absolute
base and payload positions along the selected direction, and define the
payload displacement relative to the loaded equilibrium as
\begin{align}
x &= x_p-x_b-x_0 ,
\label{eq:relative_payload_displacement}\\[-2pt]
F_{\rm int} &= k_e x+c_e\dot{x},
\label{eq:interaction_force}\\[-2pt]
m_p\ddot{x}_p &= F_{\rm ext}-F_{\rm int},
\label{eq:payload_dynamics}\\[-2pt]
m_p\ddot{x}+c_e\dot{x}+k_e x
&=F_{\rm ext}-m_p\ddot{x}_b .
\label{eq:base_excited_payload}
\end{align}
Here, $x_0$ is the nominal base-to-payload offset,
$F_{\rm int}>0$ is the force exerted by the payload on the robot-arm
subsystem, and $F_{\rm ext}>0$ is an external force applied to the
payload along the considered positive direction. 
Equation \eqref{eq:base_excited_payload} therefore captures both external-force
and locomotion-induced base excitation: $F_{\rm ext}$ directly excites
the payload, whereas base acceleration enters as the inertial forcing
term $-m_p\ddot{x}_b$. If the base is fixed, then
$\ddot{x}_b=0$ and the equation reduces to the standard
mass-spring-damper oscillator driven only by $F_{\rm ext}$.
To connect the passive arm-payload response to the \gls{zmp}-margin
analysis, we use the robot-arm free-body convention of Sec.~\ref{subsec:zmp_concept}. 
Assuming negligible payload vertical acceleration and no additional
external vertical force, $f_{p,z}\simeq -m_pg$. In the reduced
point-mass representation, the payload interaction is applied at the
payload point mass, so that $x_e\simeq x_p$ and $z_e\simeq z_p$, with
$M=m_R+m_p$. We express the payload position relative to the
robot-arm subsystem CoM as $x_p=x_R+x_0+x$, where $x_0$ is the nominal loaded offset and $x$ is the
relative payload displacement.  Equation~\eqref{eq:zmp_external} then becomes:
\begin{equation}
x_{\rm zmp}
=
x_R +\frac{m_p}{M}x_0+\frac{m_p}{M}x
+\frac{z_p}{Mg}F_{\rm int}
-\frac{m_Rz_R}{Mg}\ddot{x}_b .
\label{eq:zmp_decomposition}
\end{equation}
We define the incremental ZMP displacement by subtracting the nominal
loaded-equilibrium contribution:
\begin{align}
\Delta x_{\rm zmp}
&=x_{\rm zmp}-x_R-\frac{m_p}{M}x_0 ,
\label{eq:zmp_increment_definition}\\[-2pt]
&=\frac{m_p}{M}x
+\frac{z_p}{Mg}(k_e x+c_e\dot{x})
-\frac{m_Rz_R}{Mg}\ddot{x}_b .
\label{eq:incremental_zmp}
\end{align}
The three terms on the right-hand side of \eqref{eq:incremental_zmp} represent
separate contributions to the ZMP displacement: the first is
due to relative payload displacement, the second to the interface force
transmitted through the passive arm, and the third to robot-base
acceleration. The same relation follows from the complete
robot-arm-payload free body, where $F_{\rm int}$ cancels with the
equal-and-opposite interface force acting on the payload. Thus, payload
gravity, inertia, and interface forces are not double-counted.
Taking the Laplace transform of~\eqref{eq:base_excited_payload} under
zero initial conditions, with
$D(s)=m_ps^2+c_es+k_e$ and
$A_b(s)=\mathcal L\{\ddot{x}_b\}$, gives
\begin{align}
X(s)
&=\frac{F_{\rm ext}(s)}{D(s)}
-\frac{m_pA_b(s)}{D(s)},
\label{eq:payload_dual_input}\\[-2pt]
\Delta X_{\rm zmp}(s)
&=G_F(s)F_{\rm ext}(s)+G_a(s)A_b(s),
\label{eq:zmp_dual_input}
\end{align}
where
\begin{align}
G_F(s)
&=
\frac{m_pg+z_p(k_e+c_es)}
{Mg(m_ps^2+c_es+k_e)},
\label{eq:force_zmp_tf}\\[-2pt]
G_a(s)
&=
-\frac{m_p[m_pg+z_p(k_e+c_es)]}
{Mg(m_ps^2+c_es+k_e)}
-\frac{m_Rz_R}{Mg}.
\label{eq:base_zmp_tf}
\end{align}
Here, $G_F$ maps the external payload force $F_{\rm ext}$ to
$\Delta x_{\rm zmp}$ for a fixed base, while $G_a$ maps base
acceleration $\ddot{x}_b$ to $\Delta x_{\rm zmp}$ when
$F_{\rm ext}=0$. Since $\ddot{x}_b$ depends on the gait, $G_a$ links
locomotion-induced base motion to the ZMP response.

\begin{figure}[ht!]
\centering
\includegraphics[width=1\linewidth]{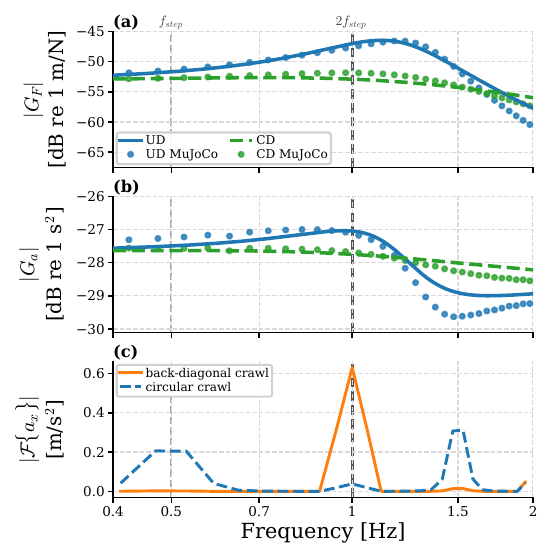}
\caption{Frequency-domain analysis of the passive payload interface:
(a) force-to-ZMP transfer function $G_F$ for \gls{ud} and \gls{cd}
configurations, with MuJoCo validation;
(b) base-acceleration-to-ZMP transfer function $G_a$ with MuJoCo
validation; and
(c) FFT magnitude of longitudinal base acceleration for back-diagonal and
circular crawl. Vertical lines indicate the $0.5\,\mathrm{Hz}$ step
frequency and its second harmonic.}
\label{fig:bode_FFT}
\end{figure}

To assess the accuracy of the reduced analytical model, we compare its
frequency response against MuJoCo \cite{Mujoco} simulations of the non-reduced
quadruped--arm rigid-body model. The analytical transfer
functions~\eqref{eq:force_zmp_tf} and~\eqref{eq:base_zmp_tf} are
validated over the nine payload/damping conditions of
Table~\ref{tab:damping_ratios} using logarithmic chirp inputs over
$0.1$--$5\,\mathrm{Hz}$. For $G_F$, the base is fixed and a force chirp
is applied at the end effector; for $G_a$, the trunk is kept fixed and
each passive-arm body is excited by the equivalent inertial force
associated with a logarithmic base-acceleration chirp, with
$F_{\rm ext}=0$. For this validation, the scalar effective moving mass is refined to include
the projected inertial contribution of the passive-arm links, rather than
using the payload-only approximation $m_e\simeq m_p$. With this refinement,
the reduced model matches the simulated force-to-ZMP response with
$1.72\,\mathrm{dB}/11.2^\circ$ RMS error and the
base-acceleration-to-ZMP response with $0.58\,\mathrm{dB}/6.1^\circ$
RMS error. Table~\ref{tab:all_paths}
summarizes the validation errors for all six input--output paths.
All paths remain within $2.29\,\mathrm{dB}$
magnitude error; the largest phase error occurs for
$\ddot{x}_b\!\rightarrow\!F_{\rm int}$ because the reduced model
neglects arm-link inertial reaction to base acceleration.

\begin{table}[!ht]
\caption{RMS errors for the six input--output validation paths.}
\centering
\footnotesize
\setlength{\tabcolsep}{3pt}
\begin{tabular}{l l c}
\toprule
Input & Output & RMS error (magnitude/phase) \\
\midrule
$F_{\rm ext}$ & ZMP & $1.72\,\mathrm{dB}/11.2^\circ$ \\
$F_{\rm ext}$ & Payload & $1.17\,\mathrm{dB}/6.3^\circ$ \\
$F_{\rm ext}$ & Interface & $1.60\,\mathrm{dB}/11.9^\circ$ \\
$\ddot{x}_b$ & ZMP & $0.58\,\mathrm{dB}/6.1^\circ$ \\
$\ddot{x}_b$ & Payload & $1.43\,\mathrm{dB}/5.9^\circ$ \\
$\ddot{x}_b$ & Interface & $2.29\,\mathrm{dB}/33.5^\circ$* \\
\bottomrule
\end{tabular}
\label{tab:all_paths}
\end{table}
\vspace{-0.3cm}
\subsection{Frequency response of the theoretical model}

Figure~\ref{fig:bode_FFT}(a)-(b) validates the two excitation paths:
external payload force mapped through $G_F$ and locomotion-induced base
acceleration through $G_a$. The \gls{ud} case shows a pronounced
resonance near $1\,\mathrm{Hz}$, while the \gls{cd} response is
attenuated. Since $G_a$ maps robot-base acceleration to \gls{zmp}
displacement, it links gait-induced base motion to the resulting
\gls{zmp} response.
For a measured base-acceleration spectrum $A_b(f)$, the
locomotion-induced \gls{zmp} contribution is estimated as
$|G_a(j2\pi f)|\,|A_b(f)|$, directly linking gait-induced base motion
to the \gls{zmp} response.
To characterize the locomotion excitation, we analyze the longitudinal
base acceleration of two crawl gaits using the fast Fourier transform
(FFT):
\textit{back-diagonal crawl}
(\gls{lf}-\gls{rh}-\gls{rf}-\gls{lh}) and \textit{circular crawl}
(\gls{lf}-\gls{rf}-\gls{rh}-\gls{lh}), where the sequences denote the
swing-leg order. Both use a step frequency of $0.5\,\mathrm{Hz}$ and a
duty factor of $0.7$, differing only in contact sequence.
The back-diagonal sequence alternates front and rear legs across the
robot diagonals, producing higher periodic longitudinal excitation. The
circular sequence instead leads to a circular-like base motion where the base
moves back and forth, along both longitudinal and lateral directions, only once
per stride.

Figure~\ref{fig:bode_FFT}(c) shows that the two gait sequences produce
different longitudinal base-acceleration spectra despite identical stride timing.
The circular crawl exhibits a dominant spectral component around
$f_{\mathrm{step}}=0.5\,\mathrm{Hz}$, whereas the back-diagonal crawl produces
a stronger dominant peak near the second harmonic
$2f_{\mathrm{step}}$ (i.e., $1\,\mathrm{Hz}$).
For the passive-arm configuration studied here, the identified
\gls{ud} resonance lies near the second harmonic of the back-diagonal
crawl base acceleration. This base-acceleration component is therefore
mapped to \gls{zmp} motion through $G_a$. This shows how locomotion alone can excite the passive payload-interface
mode and reduce the ZMP margin when the passive resonance lies near a
gait-induced excitation frequency. Increasing damping reduces this sensitivity, motivating the
damping-regime analysis in Sec.~\ref{sec:section_violins}.

\section{Simulation Analysis of Damping Conditions}
\label{sec:section_violins}
The simulations and hardware validation use the PACC platform
~\cite{PACC_Paper}: a $23\,\mathrm{kg}$ Unitree Aliengo equipped with a
$1.5\,\mathrm{kg}$ three-joint passive arm with encoder-measured joint
positions. The external wrench is estimated using the unchanged momentum
observer of~\cite{PACC_Paper}.
All simulations are performed in MuJoCo using a back-diagonal crawl
(\gls{lf}-\gls{rh}-\gls{rf}-\gls{lh}) at $0.1\,\mathrm{m/s}$, selected
because Sec.~\ref{sec:impedance_and_stability} showed stronger longitudinal
base excitation. To isolate passive-arm damping effects, we use the
\gls{mpc} of~\cite{PACC_Paper} and vary only the arm damping among
\gls{ud}, \gls{cd}, and \gls{od}. The \gls{zmp} is softly constrained
inside the support polygon with a $0.04\,\mathrm{m}$ margin.

\begin{table}[!ht]
\caption{Natural frequency and damping ratio of the distal joint-3 mode.}
\centering
\footnotesize
\setlength{\tabcolsep}{3pt}
\resizebox{\columnwidth}{!}{%
\begin{tabular}{l c c c}
\hline
Payload & UD ($f_n$ [Hz], $\zeta$) &
CD ($f_n$ [Hz], $\zeta$) &
OD ($f_n$ [Hz], $\zeta$) \\
\hline
2.5\,kg & 1.332,\ 0.027 & 1.332,\ 1.016 & 1.332,\ 4.003 \\
5\,kg   & 1.174,\ 0.153 & 1.174,\ 0.996 & 1.174,\ 2.271 \\
8\,kg   & 1.109,\ 0.290 & 1.109,\ 1.082 & 1.109,\ 2.904 \\
\hline
\end{tabular}%
}
\label{tab:damping_ratios}
\end{table}

\begin{figure}
    \centering
    \includegraphics[width=1\linewidth]{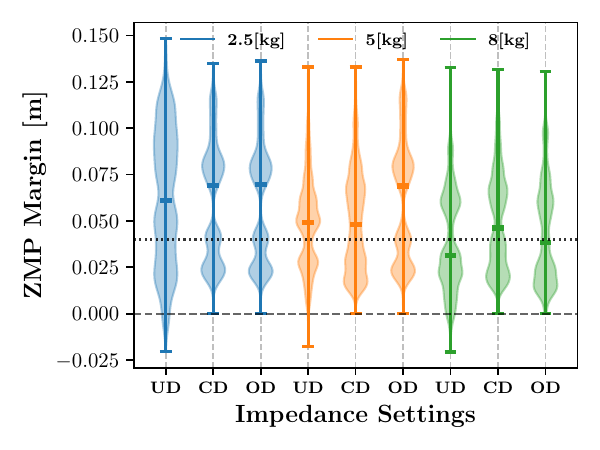}
   \caption{\gls{zmp}-margin distributions for $2.5$, $5$, and $8\,\mathrm{kg}$ payloads under \gls{ud}, \gls{cd}, and \gls{od} passive-arm configurations on flat terrain. The dotted horizontal line indicates the desired $0.04\,\mathrm{m}$ safety margin.}
    \label{fig:violin_plot_flat_single}
\end{figure}

As reported in Table~\ref{tab:damping_ratios}, the natural frequencies
range from approximately $1.11$ to $1.33\,\mathrm{Hz}$; the \gls{cd} settings
give $\zeta\simeq1$, while the \gls{ud} and \gls{od} settings lie below and above
critical damping, respectively.
Figure~\ref{fig:violin_plot_flat_single} shows the \gls{zmp}-margin
distributions using violin plots. Each violin summarizes a
$40\,\mathrm{s}$ MuJoCo trajectory for one payload and damping condition,
corresponding to approximately 20 gait cycles, and the lower tails
highlight margin-violation behavior not captured by the mean.
At $2.5\,\mathrm{kg}$, all configurations remain mostly above the desired $0.04\,\mathrm{m}$ margin. At $5$ and $8\,\mathrm{kg}$, the \gls{ud} case develops wider distributions and deeper lower tails, whereas the \gls{cd} case remains more compact; the \gls{od} configuration also limits oscillations but exhibits greater margin dispersion at higher payloads. These trends are consistent with Sec.~\ref{sec:impedance_and_stability}: the \gls{ud} configuration exhibits a pronounced resonance near $1\,\mathrm{Hz}$, where the crawl gait also contains significant excitation. The resulting passive-arm oscillations therefore produce larger \gls{zmp}-margin variations, while the \gls{cd} configuration attenuates this resonant response. 
\looseness=-1
\section{Extended SRBD MPC Formulation}
\label{sec:arMPC}

The previous sections showed that locomotion can excite the passive arm
and reduce the available \gls{zmp} margin. We therefore formulate
\gls{armpc} as an extended \gls{srbd}-based \gls{mpc} that augments the
reduced floating-base prediction model with passive-arm dynamics. The
formulation retains the reduced floating-base representation used in
standard \gls{srbd}-based \gls{mpc}, while adding the passive-arm joint
states, internal spring-damper dynamics, and base--arm inertial
coupling. The formulation is detailed next and evaluated in simulation
and hardware.
\vspace{-0.2cm}
\subsection{Extended SRBD-Passive-Arm Dynamics}

We consider the quadruped robot equipped with the passive arm shown in
Fig.~\ref{fig:Arm_views}. In the proposed prediction model, the
quadruped legs are not modeled explicitly, as in common \gls{srbd}-based
quadruped \gls{mpc} formulations; their effect enters through the ground
reaction forces applied at the contact feet. The passive arm is retained
through its joint states, rigid-body dynamics, torsional springs and
dampers, and inertial coupling with the base. The resulting model
predicts both arm-induced base effects and base-induced arm motion.
Let $n_a$ denote the number of passive joints ($n_a=3$ in the
implementation used in this work). The coupled dynamics are written as
\begingroup
\setlength{\arraycolsep}{2pt}
\renewcommand{\arraystretch}{0.95}
\begin{equation}
\begin{aligned}
&
\begin{bmatrix}
\mathbf M_b & \mathbf M_{ba} \\
\mathbf M_{ab} & \mathbf M_a
\end{bmatrix}
\begin{bmatrix}
\dot{\boldsymbol{\nu}}_b \\
\ddot{\mathbf q}_a
\end{bmatrix}
+
\begin{bmatrix}
\mathbf h_b \\
\mathbf h_a
\end{bmatrix}
+
\begin{bmatrix}
\mathbf 0_{6\times6} & \mathbf 0_{6\times n_a} \\
\mathbf 0_{n_a\times6} & \mathbf K_a
\end{bmatrix}
\begin{bmatrix}
\mathbf q_b \\
\mathbf q_a-\mathbf q_{a,0}
\end{bmatrix}
\\[-1pt]
&
+
\begin{bmatrix}
\mathbf 0_{6\times6} & \mathbf 0_{6\times n_a} \\
\mathbf 0_{n_a\times6} & \mathbf D_a
\end{bmatrix}
\begin{bmatrix}
\boldsymbol{\nu}_b \\
\dot{\mathbf q}_a
\end{bmatrix}
=
\begin{bmatrix}
\mathbf w \\
\mathbf 0_{n_a}
\end{bmatrix}.
\end{aligned}
\label{eq:augmented_dynamics}
\end{equation}
\endgroup
where $\mathbf q_b$ and $\boldsymbol{\nu}_b$ denote the floating-base
pose and spatial velocity, while $\mathbf q_a$ and $\dot{\mathbf q}_a$
denote the passive-arm joint positions and velocities, consistently
with the notation introduced in Sec.~\ref{subsec:eq_model_passive_arm}.
Here,
$\mathbf M_b\in\mathbb{R}^{6\times6}$,
$\mathbf M_a\in\mathbb{R}^{n_a\times n_a}$,
$\mathbf M_{ba}\in\mathbb{R}^{6\times n_a}$, and
$\mathbf M_{ab}\in\mathbb{R}^{n_a\times6}$, while
$\mathbf K_a,\mathbf D_a\in\mathbb{R}^{n_a\times n_a}$ contain the
arm-joint stiffness and damping coefficients; $\mathbf q_{a,0}$ denotes the passive-arm rest configuration.
The passive-arm inertia block $\mathbf M_a$, the coupling blocks
$\mathbf M_{ba}$ and $\mathbf M_{ab}=\mathbf M_{ba}^\top$, and the
passive bias term $\mathbf h_a$ are extracted from the corresponding rows and columns of
the full rigid-body model using composite-rigid-body inertia and
bias-force computations implemented with Pinocchio~\cite{pinocchioweb}.
These quantities are recomputed at every \gls{mpc} update from the current
estimated robot-arm configuration and are then evaluated along the
prediction horizon using the predicted passive-arm states.
The floating-base block $\mathbf M_b$ and bias term $\mathbf h_b$ are
obtained from the corresponding rows and columns of the complete
robot-arm inertia matrix, with the leg joints fixed in a nominal standing configuration:
$\mathbf M_b =
\mathbf M_{\mathrm{full}}(\mathbf q)\big|_{b,b}$.
Similarly, $\mathbf h_b$ is obtained from the corresponding floating-base
components of the full bias-force vector, including gravity and
rotational centrifugal terms, under the same nominal leg configuration.
\looseness=-1

The wrench generated by the ground reaction forces is
\begin{equation}
\mathbf w =
\begin{bmatrix}
\displaystyle\sum_i c_i\mathbf f_i \\
\displaystyle\sum_i c_i\,\mathbf p_i\times\mathbf f_i
\end{bmatrix}
\label{eq:grf_wrench}
\end{equation}
where $\mathbf f_i\in\mathbb{R}^3$ is the \gls{grf} at foot $i$,
$\mathbf p_i$ is the vector from the robot \gls{com} to the
corresponding contact point, and $c_i\in\{0,1\}$ is the contact
indicator provided by the predefined gait schedule.

The coupled system in~\eqref{eq:augmented_dynamics} is resolved at each
prediction node through a Schur-complement reduction of the passive-arm
block. No measured or observer-estimated end-effector force is required
inside \gls{armpc}'s prediction; the arm-base interaction follows
directly from the coupled dynamics.

\begingroup
\begingroup
\setlength{\abovedisplayskip}{6pt}
\setlength{\belowdisplayskip}{5pt}
\setlength{\abovedisplayshortskip}{4pt}
\setlength{\belowdisplayshortskip}{4pt}
\subsection{Optimal Control Problem}

The \gls{armpc} state augments the standard \gls{srbd} state with the
passive arm joint positions and velocities:
\begin{equation}
\mathbf x =
\begin{bmatrix}
\mathbf p_c^\top,\,
\mathbf v_c^\top,\,
\boldsymbol{\eta}^\top,\,
\boldsymbol{\omega}_B^\top,\,
\mathbf q_a^\top,\,
\dot{\mathbf q}_a^\top
\end{bmatrix}^{\top}
\label{eq:armpc_state}
\end{equation}
where $\mathbf p_c,\mathbf v_c\in\mathbb{R}^3$ are the CoM position and
velocity, $\boldsymbol{\eta},\boldsymbol{\omega}_B\in\mathbb{R}^3$ are
the base orientation and angular velocity, and
$\mathbf q_a,\dot{\mathbf q}_a\in\mathbb{R}^{n_a}$ are the passive-arm
joint states. For the three-joint arm used here, $\mathbf x\in\mathbb{R}^{18}$.

The control inputs are the four
foot \glspl{grf},
\begin{equation}
\mathbf u =
\begin{bmatrix}
\mathbf f_{\rm FL}^\top &
\mathbf f_{\rm FR}^\top &
\mathbf f_{\rm RL}^\top &
\mathbf f_{\rm RR}^\top
\end{bmatrix}^{\top}
\in\mathbb{R}^{12}
\label{eq:armpc_input}
\end{equation}
The contact schedule is prescribed through $c_{i,k}$, with only stance
forces contributing to~\eqref{eq:grf_wrench} and zero reference
\gls{grf} assigned to swing legs. The dynamics
in~\eqref{eq:augmented_dynamics} are integrated with explicit
Runge--Kutta using $\Delta t=0.02\,\mathrm{s}$ and $N=25$, yielding a
$0.5\,\mathrm{s}$ prediction horizon.

The \gls{ocp} minimizes the cost:

\begin{equation}
\begin{aligned}
J={}&
\sum_{k=0}^{N-1}
\left(
\|\mathbf x_k-\mathbf x_k^{\rm ref}\|_{\mathbf Q}^{2}
+\|\mathbf u_k-\mathbf u_k^{\rm ref}\|_{\mathbf R}^{2}
\right)\\[-2pt]
&+\|\mathbf x_N-\mathbf x_N^{\rm ref}\|_{\mathbf Q_N}^{2}
\end{aligned}
\label{eq:armpc_cost}
\end{equation}

where $\mathbf Q$, $\mathbf Q_N$, and $\mathbf R$ are diagonal weighting
matrices. In addition to the standard \gls{srbd} tracking terms, the
cost penalizes passive-arm displacement and velocity through
$\|\mathbf q_a-\mathbf q_{a,0}\|_{\mathbf Q_{q_a}}^2+
\|\dot{\mathbf q}_a\|_{\mathbf Q_{\dot q_a}}^2$.
In Sec.~\ref{subsec:eef_tracking_task}, to perform an end-effector tracking task, a quadratic tracking term for the end-effector position is added to the cost. 
For each stance contact, the \glspl{grf} satisfy the inequality constraints
\begin{equation*}
|f_{x,i}|,\,|f_{y,i}|\leq \mu f_{z,i},\qquad
0\leq f_{z,i}\leq mg,\qquad
\mu=0.5 .
\end{equation*}
where $m$ is the total modeled mass of the robot-arm system.
Locomotion stability is enforced using the \gls{zmp} formulation of
Sec.~\ref{sec:impedance_and_stability}. The six support-polygon
inequalities are collected in
$\mathbf h_{\rm ZMP}(\mathbf x_k,\mathbf u_k)\in\mathbb{R}^{6}$ and
softened as
\begin{equation}
\underline{\mathbf h}-\mathbf s_{l,k}
\leq\mathbf h_{\rm ZMP}(\mathbf x_k,\mathbf u_k)
\leq\overline{\mathbf h}+\mathbf s_{u,k},
\quad
\mathbf s_{l,k},\mathbf s_{u,k}\geq\mathbf 0
\label{eq:zmp_soft_constraint}
\end{equation}
where the bounds define the support polygon reduced by
$d_{\mathrm{ZMP}}=0.04\,\mathrm{m}$. 

The six ZMP inequalities are softened at the intermediate shooting nodes
using linear slack weights $z_l=500$ and $z_u=1000$ and quadratic
weights $Z_l=Z_u=1$. The same penalties are used for all controllers.
The OCP is solved with acados~\cite{acados} using a Gauss--Newton Hessian
and HPIPM~\cite{HPIPM}. At each \gls{mpc} update, $\mathbf{x}_0=\hat{\mathbf{x}}(t)$ and the
ZMP constraints are softened. One SQP iteration with at most
20 QP iterations is performed, warm-started from the previous
solution. The first optimized GRF is applied as
$\boldsymbol{\tau}_{i}^{\rm GRF}=-\mathbf J_i^\top\mathbf f_{i,0}^{\rm MPC}$.

\begin{figure}[!ht]
    \centering
    \includegraphics[width=1\linewidth]{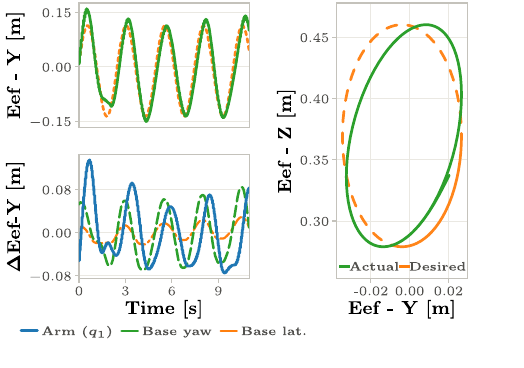}
\caption{Passive-arm end-effector tracking. \textbf{Left}: hardware
tracking of a $0.4\,\mathrm{Hz}$ sinusoidal reference along the lateral
($y$) direction, with decomposition into passive-arm, base-yaw, and
base-translation contributions. \textbf{Right}: circular end-effector
tracking in the $y$-$z$ plane in simulation. Solid and dashed lines
denote measured and desired trajectories, respectively.}
    \label{fig:eef_tracking_sim}
\end{figure}
\vspace{-0.2cm}
\subsection{Ablation study of the MPC}

We compare \gls{armpc} against two \gls{mpc} variants to isolate the
effect of passive-arm modeling on the \gls{zmp}-margin behavior.
\textit{Nominal \gls{mpc}} is the standard \gls{srbd}-based controller, where
the passive arm is not included in the prediction model.
\textit{Baseline \gls{mpc}} denotes the controller used in~\cite{PACC_Paper}.
It uses an \gls{srbd} prediction model augmented with a disturbance wrench
$\mathbf{d}\in\mathbb{R}^6$ initialized from the momentum-observer
estimate. As done in \cite{PACC_Paper}, the longitudinal and lateral
forces and the yaw moment are propagated as a spring response to base
motion relative to a fixed end-effector position, while
the vertical force and the roll and pitch moments are held constant
over the horizon. \textit{\gls{armpc}} is the proposed
\gls{srbd}--passive-arm \gls{mpc}, where passive-arm joint states,
torsional springs and dampers, and base-arm inertial coupling are
included in the prediction dynamics. The \gls{mpc} weights are identical
for all three variants and reported in
Table~\ref{tab:mpc_weights_ablation}.
\looseness=-1
\begin{table}[!h]
\caption{MPC weights used in the ablation study.
Vectors follow $(x,y,z)$ for $Q_p,Q_v,Q_f,R_{\rm grf}$ and
$(r,p,y)$ for $Q_\theta,Q_\omega$.}
\label{tab:mpc_weights_ablation}
\centering
\footnotesize
\setlength{\tabcolsep}{2.5pt}
\begin{tabular}{@{}lc lc@{}}
\toprule
Weight & Value & Weight & Value \\
\midrule
$Q_p$      & $(0,0,1500)$
& $Q_{q_a}$   & $(5,5,5)$ \\

$Q_v$      & $(200,200,200)$
& $Q_{\dot q_a}$ & $(10,10,10)$ \\

$Q_\theta$ & $(200,200,0)$
& $Q_f$     & $(300,300,300)$ \\

$Q_\omega$ & $(20,20,50)$
& $R_{\rm grf}$ & $(10^{-3},10^{-3},10^{-3})$ \\
\bottomrule
\end{tabular}
\end{table}

All variants are tested in MuJoCo with a $2\,\mathrm{kg}$ payload while
performing a back-diagonal crawl. The passive arm uses the stiffness and
damping values measured on the real robot,
$\mathbf k=[5,8,5]^\top\,\mathrm{Nm/rad}$ and
$\mathbf d=[0.5,0.5,0.5]^\top\,\mathrm{Nms/rad}$.

Three scenarios are tested: undisturbed flat-terrain locomotion;
repeated $30\,\mathrm{N}$ end-effector impulses
($0.2\,\mathrm{s}$ duration, $2\,\mathrm{s}$ spacing); and a
$15\,\mathrm{N}$ sinusoidal end-effector disturbance at $1\,\mathrm{Hz}$.
The disturbances are not provided to the controllers as known inputs.
\looseness=-1
Each controller--scenario pair is repeated over 10 matched randomized
seeds. Across runs, the ground friction coefficient is sampled from
$\mathcal{U}(1.2,1.8)$, the forward velocity from
$\mathcal{U}(0.08,0.10)\,\mathrm{m/s}$, the disturbance onset from
$\mathcal{U}(0,0.5)\,\mathrm{s}$ relative to the gait phase, and the
disturbance direction within a $20^\circ$ cone around the world-frame
$-x$ direction. Performance is evaluated using the percentage of
\gls{zmp}-margin violations, base-orientation \gls{rms} errors, and
vertical base-motion metrics. 
The \gls{zmp}-margin violation percentage is defined as the percentage
of trajectory samples for which the \gls{zmp} margin falls below the
desired $0.04\,\mathrm{m}$ threshold. It therefore measures how often
the controller loses the prescribed stability margin, while the
orientation and height metrics quantify the resulting postural response.

\begin{table}[t]
\caption{Simulation ablation under a $2\,\mathrm{kg}$ payload over
10 randomized seeds per condition. Violation percentage is
mean $\pm$ standard deviation; RMS metrics are averaged over completed runs.}
\label{tab:controller_comparison}
\vspace{-0.12cm}
\centering
\footnotesize
\setlength{\tabcolsep}{1.6pt}
\renewcommand{\arraystretch}{0.95}

\begin{tabular}{@{}l l c c c c c@{}}
\toprule
Scen. & Controller &
\shortstack{Viol.\\(\%)} &
\shortstack{Pitch\\RMS [deg]} &
\shortstack{Roll\\RMS [deg]} &
\shortstack{$z$\\RMS [mm]} &
Fail \\
\midrule

Flat
& Nominal
& $44.0\pm1.2$
& $16.8$
& $7.0$
& $14.6$
& $0/10$ \\

& \emph{Baseline}
& $9.5\pm0.8$
& $1.7$
& $\mathbf{1.7}$
& $14.7$
& $0/10$ \\

& \textbf{ARMPC}
& $\mathbf{3.3\pm0.1}$
& $\mathbf{1.4}$
& $1.9$
& $\mathbf{6.7}$
& $\mathbf{0/10}$ \\

\midrule

Impulse
& Nominal
& -- & -- & -- & -- & $10/10$ \\

& \emph{Baseline}
& $22.7\pm6.3$
& $8.3$
& $\mathbf{4.5}$
& $12.2$
& $4/10$ \\

& \textbf{ARMPC}
& $\mathbf{13.3\pm3.3}$
& $\mathbf{6.8}$
& $5.3$
& $\mathbf{10.8}$
& $\mathbf{0/10}$ \\

\midrule

Sin.
& Nominal
& $44.1\pm2.5$
& $16.9$
& $7.7$
& $13.7$
& $5/10$ \\

& \emph{Baseline}
& $18.7\pm3.7$
& $4.1$
& $2.9$
& $16.8$
& $2/10$ \\

& \textbf{ARMPC}
& $\mathbf{4.1\pm1.6}$
& $\mathbf{2.4}$
& $\mathbf{2.2}$
& $\mathbf{7.6}$
& $\mathbf{0/10}$ \\

\bottomrule
\end{tabular}
\vspace{-0.15cm}
\end{table}

Table~\ref{tab:controller_comparison} shows that explicitly predicting
the passive-arm dynamics improves robustness across the tested
conditions. Across all scenarios, \gls{armpc} gives the lowest
percentage of samples violating the prescribed \gls{zmp} margin and
completes all runs without failure.
The Nominal controller fails in all impulse trials and in 5/10 sinusoidal
trials, while \emph{Baseline} still fails in 4/10 and 2/10 runs, respectively.

The flat-terrain case confirms that the passive arm can affect stability
even without external disturbances, since the lightly damped passive mode
lies close to the crawl frequency and is excited by locomotion itself.
Baseline reduces the violation percentage through its
spring-based disturbance propagation, but \gls{armpc} further reduces it from
$9.5\%$ to $3.3\%$ by predicting passive-arm motion over the horizon,
while also reducing the vertical base \gls{rms} motion from
$14.7$ to $6.7\,\mathrm{mm}$.

Under impulse and sinusoidal disturbances, \gls{armpc} remains the
only controller with no failures and gives the lowest \gls{zmp}-margin
violation percentage.
Overall, the extended \gls{srbd}-passive-arm prediction model improves
both stability margin and robustness across the tested conditions.
\vspace{-0.4cm}
    
\subsection{Passive-Arm End-Effector Tracking}
\label{subsec:eef_tracking_task}
Since the passive-arm joint positions are included in the \gls{mpc} state and
the whole-body passive-arm model is retained in the dynamics, the
end-effector position can be evaluated along the prediction horizon
through forward kinematics. Cartesian end-effector tracking is obtained
by adding
\begin{equation}
J_{\rm ee}
=
\sum_{k=1}^{N}
\left\|
\mathbf{fk}_{\rm ee}(\mathbf x_k)
-
\mathbf p_{\rm ee}^{\rm ref}
\right\|^2_{\mathbf Q_{\rm ee}}
\label{eq:ee_tracking_cost}
\end{equation}
to the \gls{armpc} cost, where
$\mathbf{fk}_{\rm ee}(\mathbf x_k)$ denotes the passive-arm
end-effector position computed from the predicted state.

For this task, the nominal \gls{armpc} weights of
Table~\ref{tab:mpc_weights_ablation} are modified to allow the base and
passive arm to contribute to the end-effector motion:
$\mathbf Q_{\rm ee}=\mathrm{diag}(5,2\times10^4,5)$,
$\mathbf Q_p=\mathrm{diag}(50,800,1000)$,
$\mathbf Q_v=\mathrm{diag}(20,80,80)$,
$\mathbf Q_\theta=\mathrm{diag}(60,140,50)$.
During the hardware experiment, the end-effector tracks a
$0.4\,\mathrm{Hz}$ sinusoidal reference with $0.1\,\mathrm{m}$
amplitude along the lateral ($y$) direction. The resulting motion is
generated primarily through the combined motion of the passive arm and
the robot base, demonstrating that \gls{armpc} can use the passive-arm configuration to realize Cartesian
tracking.
As shown in Fig.~\ref{fig:eef_tracking_sim}, the hardware lateral tracking
RMS error is $26.6\,\mathrm{mm}$. In simulation, circular $y$-$z$ tracking reaches RMS
errors of $0.01\,\mathrm{m}$ and $0.04\,\mathrm{m}$.

\subsection{Hardware Validation}

After simulation validation, the controller is evaluated on a $23\,\mathrm{kg}$ Unitree Aliengo equipped with the passive arm as shown in Fig.~\ref{fig:Arm_views}. The floating-base state is estimated from proprioceptive sensing using MUSE~\cite{MUSE}, while passive-arm joint positions are measured by encoders.
The robot performs a back-diagonal crawl in place while carrying a $2\,\mathrm{kg}$ payload and is subjected to manual pull-release disturbances along two directions: a longitudinal pull, where the response is dominated by the distal pitch joint of the passive arm, and a diagonal pull, where the arm is pulled diagonally with respect to the robot heading so that both the first and third joints are excited. To improve repeatability, the arm encoder angle and interaction force are monitored online, and the arm is released when a prescribed reference deflection is reached. 
Both controllers use the same passive-arm interface and sensing setup.
The momentum observer is used by Baseline \gls{mpc} to estimate the
end-effector wrench, whereas \gls{armpc} does not use a measured or
observer-estimated end-effector force inside the prediction horizon.
The hardware comparison isolates the difference in predictive
control formulation.
State estimates are obtained by running the MUSE state estimator at
1\,kHz for both controllers. The hardware experiments were executed on
an AMD Ryzen AI 7 (8 cores/16 threads, 30\,GB RAM). \gls{armpc} replans at 90\,Hz, corresponding to a
control period of approximately 11\,ms, compared with 100\,Hz for
Baseline MPC. The measured QP solution time was $1.66\pm0.60$\,ms for \gls{armpc} and
$1.60\pm0.66$\,ms for Baseline MPC.
For the hardware trials reported in Table~\ref{tab:trial_counts}, a
trial is counted as a failure if any part of the robot other than the
feet contacts the ground.
For the longitudinal disturbance, the distal pitch-joint deflection at
release is approximately $-0.8\,\mathrm{rad}$ with about
$20\,\mathrm{N}$ interaction force. For the diagonal
disturbance, the proximal yaw-joint deflection spans approximately
$[-1.1,0.9]\,\mathrm{rad}$ with $30$-$35\,\mathrm{N}$ interaction
force. The same release conditions are used for both controllers.
\gls{armpc} completed all trials without failure, whereas Baseline \gls{mpc} failed in
7 of 15 longitudinal trials and 8 of 15 diagonal trials.
Further Baseline \gls{mpc} testing was stopped after repeated unsafe events
to limit the risk of hardware damage. Additional hardware results are provided in the accompanying \href{https://youtu.be/92qLK9qrMsI}{video}. 
\looseness=-1
\begin{table}[t]
\centering
\caption{Pull-release trials under a $2\,\mathrm{kg}$ payload.}
\label{tab:trial_counts}
\footnotesize
\setlength{\tabcolsep}{10pt}
\begin{tabular}{@{}llccc@{}}
\toprule
Dist. & Ctrl & $N$ & Fail. & $[\%]$ \\
\midrule
Longitudinal & ARMPC (Ours) & 30 & \textbf{0} & \textbf{0\%} \\
     & \emph{Baseline}         & 15 & 7 & 46.6\% \\
Diagonal & ARMPC (Ours) & 55 & \textbf{0} & \textbf{0\%} \\
          & \emph{Baseline}         & 15 & 8 & 53.3\% \\
\bottomrule
\end{tabular}
\end{table}

\section{Conclusions} 
\label{sec:Conclusion}

This paper presented a modeling and control framework for quadruped
payload carrying with passive arms. The reduced analytical model relates
passive-interface impedance and payload mass to the \gls{zmp} response
under both external disturbances at the mechanism end-effector and locomotion-induced base excitation,
showing that passive modes near gait harmonics can reduce the
stability margin.
Motivated by this analysis, \gls{armpc} combines reduced \gls{srbd}
base dynamics with the complete passive-arm dynamics in an extended passive-arm
predictive model. Simulation ablations show lower \gls{zmp}-margin violation percentages
and no failures with \gls{armpc}, while Nominal and Baseline MPC fail under
disturbed conditions. The same formulation enables
passive-arm end-effector tracking, and hardware pull-release tests with a
$2\,\mathrm{kg}$ payload show no \gls{armpc} failures, whereas Baseline MPC
fails in 7/15 longitudinal and 8/15 diagonal trials. A limitation of the present study is that robustness to mismatch in payload
mass, joint stiffness, and damping is not evaluated explicitly. Future work
will therefore investigate parameter-mismatch sensitivity, online payload
and impedance estimation, and collaborative carrying.







\bibliographystyle{IEEEtran}
\bibliography{literature.bib}

\end{document}